\documentclass{article}
\usepackage{spconf,amsmath,graphicx}

\usepackage{multirow}
\usepackage{framed}
\usepackage{pgfplots}

\definecolor{cyan}{rgb}{0.0, 0.65, 0.85}
\definecolor{magenta}{rgb}{0.85, 0.15, 0.5}
\definecolor{orange}{rgb}{1.0, 0.5, 0.0}

\graphicspath{{./fig/}}
\DeclareGraphicsExtensions{.eps}

\title{Character-Level incremental speech recognition \\ with recurrent neural networks}
\name{Kyuyeon Hwang and Wonyong Sung\thanks{This work was supported in part by the Brain Korea 21 Plus Project and the National Research Foundation of Korea (NRF) grant funded by the Korea government (MSIP) (No.~2015R1A2A1A10056051).}}
\address{Department of Electrical and Computer Engineering \\
Seoul National University \\
1, Gwanak-ro, Gwanak-gu, Seoul, 08826 Korea \\
\texttt{kyuyeon.hwang@gmail.com; wysung@snu.ac.kr}}
\begin{document}
\ninept
\maketitle
\begin{abstract}
In real-time speech recognition applications, the latency is an important issue.  We have developed a character-level incremental speech recognition (ISR) system that responds quickly even during the speech, where the hypotheses are gradually improved while the speaking proceeds.  The algorithm employs a speech-to-character unidirectional recurrent neural network (RNN), which is end-to-end trained with connectionist temporal classification (CTC), and an RNN-based character-level language model (LM). The output values of the CTC-trained RNN are character-level probabilities, which are processed by beam search decoding. The RNN LM augments the decoding by providing long-term dependency information. We propose tree-based online beam search with additional depth-pruning, which enables the system to process infinitely long input speech with low latency. This system not only responds quickly on speech but also can dictate out-of-vocabulary (OOV) words according to pronunciation. The proposed model achieves the word error rate (WER) of 8.90\% on the Wall Street Journal (WSJ) Nov'92 20K evaluation set when trained on the WSJ SI-284 training set.
\end{abstract}
\section{Introduction}
\label{sec:intro}

Incremental speech recognition (ISR) allows a speech-based interaction system to react quickly while the utterance is being spoken. Unlike offline sentence-wise automatic speech recognition (ASR), where the decoding result is available after a user finishes speaking, ISR returns $N$-best decoding results with small latency during speech. These $N$-best results, or hypotheses, gradually improve as the system receives more speech data. Since ISR is usually employed for immediate reaction to speech, word stability \cite{selfridge2011stability, mcgraw2011estimating} and incremental lattice generation \cite{sagerer1996incremental} have been important topics.

In this paper, we introduce an end-to-end character-level ISR system with two unidirectional recurrent neural networks (RNNs). An acoustic RNN roughly dictates the input speech and an RNN-based language model is employed to augment the dictation result through decoding. Compared to a conventional word-level backend for speech recognition system, the character-level ASR is capable of dictating out of vocabulary (OOV) words based on the pronunciation. Also, our model is trained directly from speech and text corpus and does not require external word dictionary or senone modeling.

There have been efforts to deal with OOV words in conventional HMM based ASR systems. In \cite{killer2003grapheme}, graphemes are employed as basic units instead of phonemes. Also, a sub-lexical language model is proposed in \cite{bisani2005open} for detecting previously unseen words.

RNN-based character-level end-to-end ASR systems were studied in \cite{graves2014towards, hannun2014deepspeech, hannun2014first, miao2015eesen, bahdanau2015end}. However, they lack the capability of dictating OOV words since the decoding is performed with word-level LMs. Recently, a lexicon-free end-to-end ASR system is introduced in \cite{maas2015lexicon}, where a character-level RNN LM is employed. We further improve this approach by employing prefix tree based online beam search with additional depth-pruning for ISR.

The character-level ISR system proposed in this paper is composed of an acoustic RNN and an RNN LM. The acoustic RNN is end-to-end trained with connectionist temporal classification (CTC) \cite{graves2006connectionist} using Wall Street Journal (WSJ) speech corpus \cite{paul1992design}. The output of the acoustic RNN is the probability of characters, which are decoded with character-level beam search to generate $N$-best hypotheses. To improve the performance, a character-level RNN LM is employed to augment the beam search performance. Also, we propose depth-pruning for efficient tree-based beam search. The RNN LM is separately trained with large text corpus that is also included in WSJ corpus. Unlike for word-level language modeling, conventional statistical LMs such as $n$-gram back-off models cannot be used because much longer history window is required for character-level prediction. Both acoustic RNN and RNN LM have deep unidirectional long short-term memory (LSTM) network structures \cite{hochreiter1997long, graves2013hybrid}. For continuous ISR on infinitely long input speech, they are trained with virtually infinite training data streams that are generated by randomly concatenating training sequences.

The proposed model is evaluated on a single test sequence that is generated by concatenating all test utterances in WSJ \texttt{eval92} (Nov'92 20k evaluation set) without any external reset of RNN states at the utterance boundaries. The ISR performance is examined by varying the beam width and depth. Generally, wider beam increases the accuracy. Under the same beam width, there is a trade-off between the accuracy and stability (or latency), where the balance between them can be adjusted by the beam depth.


\section{Models}
\label{sec:model}

\subsection{Acoustic model}
\label{ssec:speech}

The acoustic model is a deep RNN trained with CTC \cite{graves2006connectionist}. 
The network consists of two LSTM layers with 768 cells each, where the network has total 12.2 M trainable parameters. The model is similar to the one in the previous work about end-to-end speech recognition with RNNs \cite{graves2014towards} except a few major differences. In our case, the RNN is trained by online CTC \cite{hwang2015online} with very long training sequences that are generated by randomly concatenating several utterances. There is no need to reset the RNN states at the utterance boundary. This is necessary for ISR systems that runs continuously with an infinite input audio stream. Also, our model has a unidirectional structure since bidirectional networks that are usually employed for end-to-end speech recognition are not suitable for low-latency speech recognition. This is because the backward layers in the bidirectional networks cannot be computed before the input utterance is finished.

The input of the network is a 40-dimensional log mel-frequency filterbank feature vector with energy and their delta and double-delta values, resulting in an 123-dimensional vector. The feature vectors are extracted every 10 ms with 25 ms Hamming window. The input vectors are element-wisely standardized based on the statistics obtained from the training set. The output is a 31-dimensional vector that consists of the probabilities of 26 upper case alphabets, 3 special characters, the end-of-sentence (EOS) symbol, and the CTC blank label.

The networks are trained with stochastic gradient descent (SGD) with 8 parallel input streams on a GPU \cite{hwang2015single}. The networks are unrolled 2048 times and weight updates are performed every 1024 forward steps. The network performances are evaluated at every 10 M training frames. The evaluation is performed on total 2 M frames from the development set. The learning rate starts from $1 \times 10^{-5}$ and is reduced by the factor of $10$ whenever the WER on the development set is not improved for 6 consecutive evaluations. The training ends when the learning rate drops below $1 \times 10^{-7}$.

We trained the networks on two training sets. The first one is the standard WSJ \texttt{SI-284} set and the second one, \texttt{SI-ALL}, is the set of all speaker independent training utterances in the WSJ corpus. Note that the utterances with verbalized punctuations are removed from both training sets. Also, odd transcriptions are filtered out, which makes the final \texttt{SI-284} and \texttt{SI-ALL} sets contain roughly 71 and 167 hours of speech, respectively. WSJ \texttt{dev93} (Nov'93 20k development set) and \texttt{eval92} (Nov'92 20k evaluation set) sets are used as the development set and the evaluation set, respectively.

\subsection{Language model}
\label{ssec:lm}

\begin{figure}[!t]
	\begin{framed}
	\texttt{\footnotesize THREE ISSUES ADVANCED MICRO OF AMERICA THE ONLY WAY TO DIVERSIFY INTO TREATING MODERN ARMIES
	\vspace{5pt}\\LOOKING AHEAD TO MR. LEYSEN WITH AN INTOLERABLE POP CUT WHEN AN ALL POWERFUL STUDENT SEEKS ITS CORE DRIVING UPJOHN STOVES
	\vspace{5pt}\\AMERICAN EXPRESS HASN'T YET SWORED PARTICULARLY WITH THE RESTRUCTURING IS A COMMITMENT TO BUY POTENTIAL BUYERS IN THE OPEN MARKET}
	\end{framed}
	\caption{Example of character-level random text generation with the RNN LM.}
	\label{fig:lmgen}
\end{figure}

An RNN language model (LM) \cite{mikolov2011extensions} is employed for the proposed ISR system since conventional statistical LMs such as $n$-gram back-off models are not suitable for character-level prediction since they cannot make use of very long history windows. Specifically, the RNN LM has a deep LSTM network structure with two LSTM layers where each of them has 512 memory cells, resulting in total 3.2 M parameters.

The input of the RNN LM is a 30-dimensional vector, where the current label (character) is one-hot encoded. The output is also a 30-dimensional vector which represents the probabilities of next labels. Although the RNN LM is trained to predict the next characters with only given the current character, the past character histories are internally stored inside the RNN and used for the prediction. It is well known that RNN LM can remember contexts for very long time steps.

As for the acoustic RNN, the RNN LM is trained on a very long text stream that is generated by attaching randomly picked sentences and inserting EOS labels between sentences. The RNN LM is trained with AdaDelta \cite{zeiler2012adadelta} based SGD method for accelerated training and better annealing. The WSJ LM training text with non-verbalized punctuation, which contains about 215 M characters, is used for training the RNN LM. Randomly selected 1\% of the corpus is reserved for evaluation, on which the final bits-per-character (BPC) of the RNN LM is 1.167 (character-level perplexity of 2.245).

Random sentences can be generated following the method described in \cite{sutskever2011generating}. Briefly, the next label is randomly picked following the probabilities of the current output of the RNN LM and fed back to the RNN in the next step. By iterating these steps, texts can be sequentially generated as shown in \figurename~\ref{fig:lmgen}. From the example, it is clear that the RNN LM learned the linguistic structures as well as spellings of words that frequently appear.

\section{Character-level beam search}
\label{sec:beam}

\subsection{Tree-based CTC beam search}

\begin{figure}[!t]
	\centerline
	{%
		\includegraphics[width=2.3in]{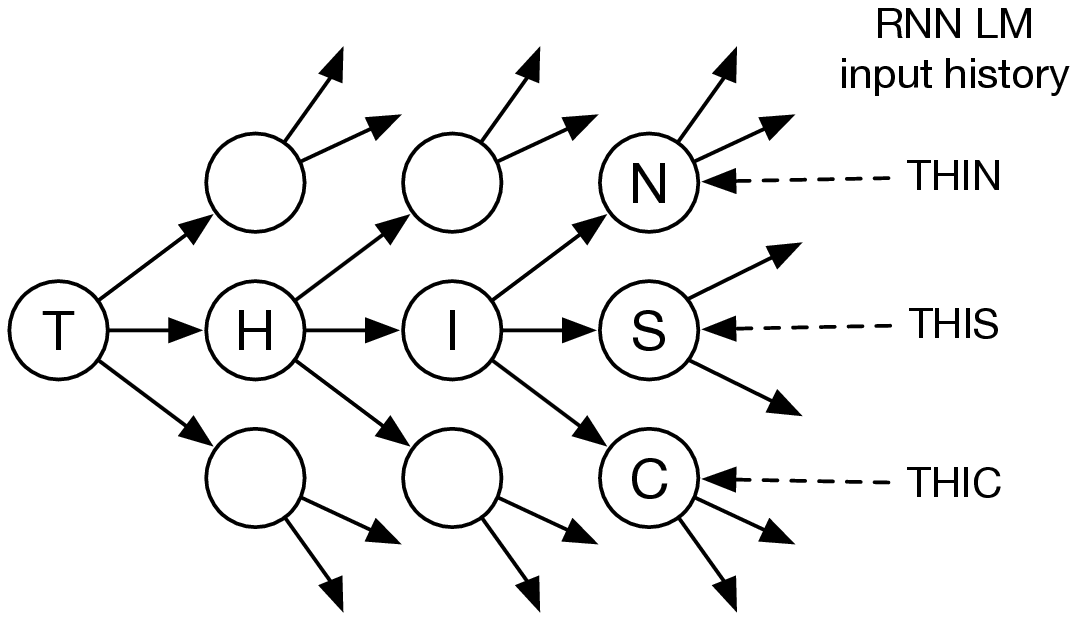}%
	}
	\caption{Beam search tree consisting of label nodes. The CTC blank label is not included.}
	\label{fig:tree}
\end{figure}

\begin{figure}[!t]
	\centerline
	{%
		\includegraphics[width=2.4in]{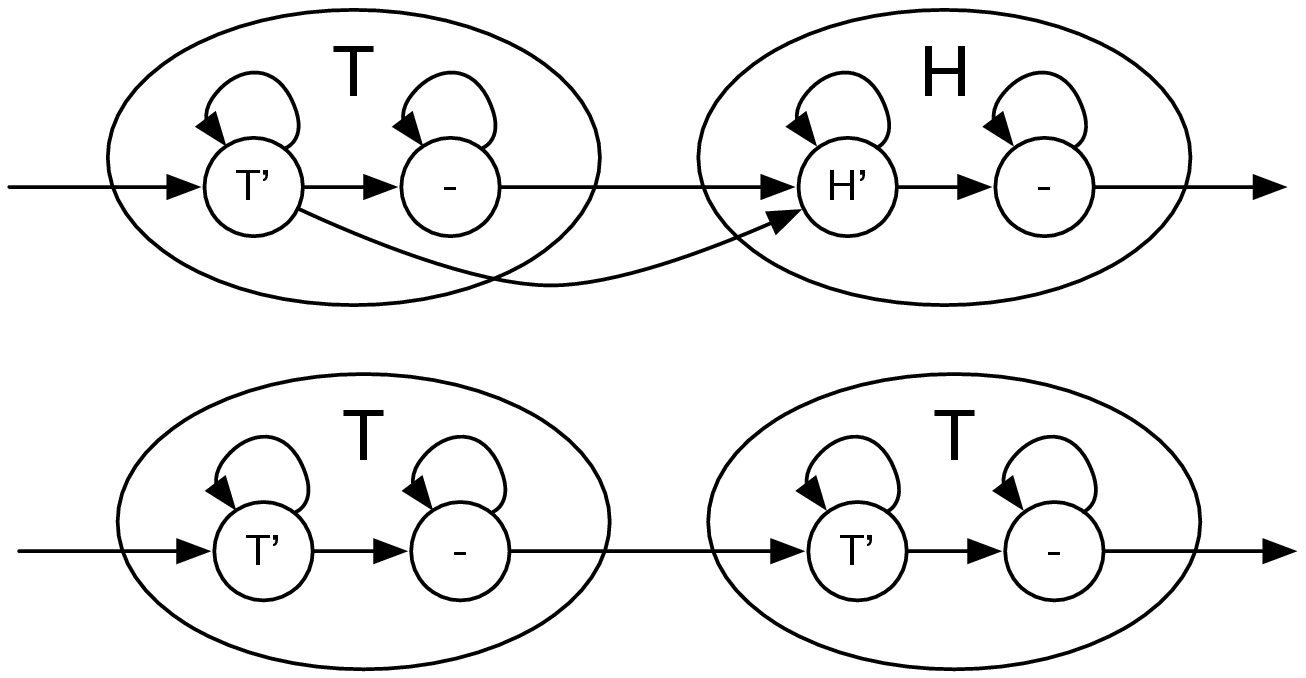}%
	}
	\caption{CTC state transition between two label nodes. If the two nodes have the same label, then a transition between the same CTC state is not allowed.}
	\label{fig:state}
\end{figure}

\begin{figure}[!t]
	\centerline
	{%
		\includegraphics[width=3.4in]{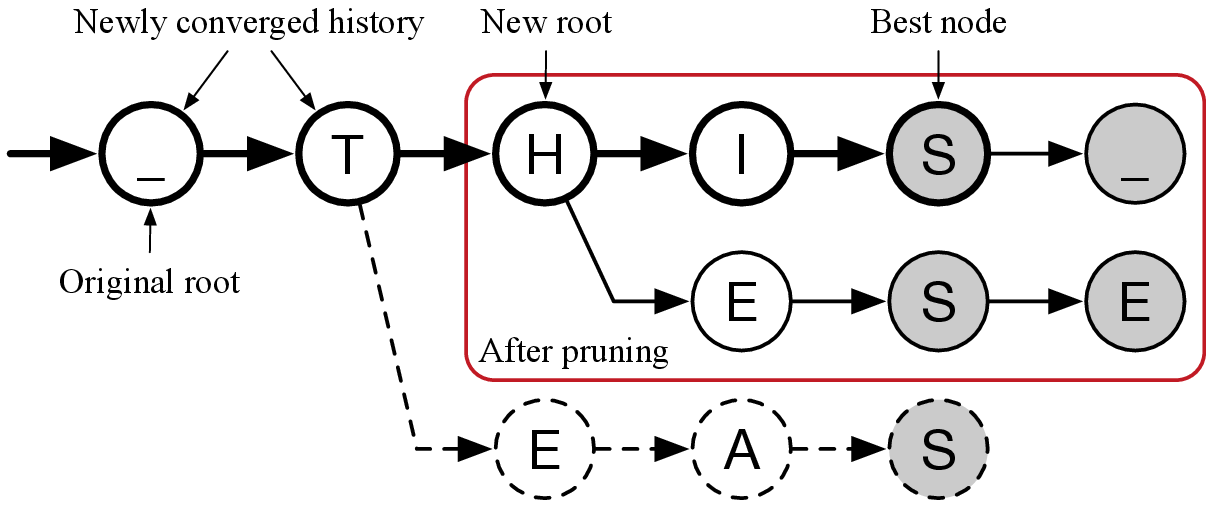}%
	}
	\caption{Example of depth-pruning with the beam depth of 2. The pruning is performed by selecting a new root node so that the new depth of the best hypothesis node becomes the beam depth. The shaded nodes indicate the original active nodes. Also, the path of the best hypothesis is drawn with thick strokes.}
	\label{fig:pruning}
\end{figure}

Let $L$ be the set of labels without the CTC blank label. The label sequence $\mathbf{z}$ is a sequence of labels in $L$. The length of the label sequence $\mathbf{z}$ is less than or equal to the number of input frames. The objective of the beam search decoding is to find the label sequence that has the maximum posterior probability given the input features from time 1 to $t$ generated by the acoustic RNNs, that is,
\begin{align}
\mathbf{z}_{\max}=\arg\max_{\mathbf{z}} P(\mathbf{z}|x_{1:t}), \label{eq:argmax_pi}
\end{align}
where $x_{1:t}$ is the input features from time 1 to $t$.

However, the CTC-trained RNN output has one more blank label. Let $L'$ be the set of labels (or CTC states) with the additional CTC blank label, and the path ${\pi}_{t}^{(i)}$ be a sequence of labels in $L'$ from time 1 to $t$. The length of the path ${\pi}_{t}^{(i)}$ is the same as $t$. By the definition of CTC, every ${\pi}$ can be reduced into the corresponding ${\mathbf{z}}$. For example, ${\pi}$ with ``aab-c--a'' corresponds to ${\mathbf{z}}$ with ``abca'', where ``-'' is the blank label.

There can be many paths, ${\pi}_{t}^{(i)}$, that can be reduced into the same ${\mathbf{z}}$. Let $\mathcal{F}(\cdot)$ be a function that maps a path to the corresponding label sequence, that is, $\mathcal{F}({\pi}_{t}^{(i)}) = {\mathbf{z}}$, then the posterior probability in (\ref{eq:argmax_pi}) becomes,
\begin{align}
P(\mathbf{z}|x_{1:t})=\sum_{\{\forall i | \mathcal{F}({\pi}_{t}^{(i)}) = {\mathbf{z}}\}}P({\pi}_{t}^{(i)}|x_{1:t}).
\label{eq:pathagg}
\end{align}
 Therefore, if the two different paths ${\pi}_{t}^{(j)}$ and ${\pi}_{t}^{(k)}$ in the decoding network are mapped to the same ${\mathbf{z}}$, then they can be merged by summing their probabilities.
 
 For the beam search, we first represent the lattice with a tree-based structure so that each node has one of labels in $L$ as depicted in \figurename~\ref{fig:tree}. Then, backtracking from any node generates a unique label sequence ${\mathbf{z}}$. To deal with CTC state transitions, we need a state-based network that is represented with CTC states, $L'$. As shown in \figurename~\ref{fig:state}, this can be easily done by expanding each tree node, of which label is in $L$, into two CTC states, one with the corresponding label in $L'$ followed by the blank CTC label. Since the label-level ($L$) search network is based on a tree structure, two different state-level ($L'$) paths with different label sequences never meet each other. This simplifies the problem since there is no interaction between two different sequence labelings (hypotheses) and (\ref{eq:pathagg}) is the only equation that we should concern. 
 
As proposed in \cite{hannun2014first, maas2015lexicon}, external language models can be integrated by modifying the posterior probability term in (\ref{eq:argmax_pi}) into:
\begin{align}
\mathrm{log}(P(\mathbf{z}|x_{1:t})) &= \mathrm{log}(P_{\mathrm{CTC}}(\mathbf{z}|x_{1:t})) \\
&\hphantom{= } + \alpha \mathrm{log}(P_{\mathrm{LM}}(\mathbf{z})) + \beta |\mathbf{z}|, \nonumber
\end{align}
where $\alpha$ is the LM weight and $\beta$ is the insertion bonus. This modification can be applied by adding the additional terms with $\alpha$ and $\beta$ to the log probability of the destination state when a state transition between two different label nodes occurs.

The probability of the next label is computed using the RNN LM when a new active label node is added to the beam search tree. For this, the RNN LM context (hidden activations) is copied from the parent node to the child node and the RNN LM processes the new label of the child node with the copied context. Therefore, each active node has its own RNN LM context.

\subsection{Pruning}

Pruning of the search tree is performed by the standard beam search approach. That is, at each frame, only the active nodes with the top $N$ hypotheses and their ancestor nodes remain alive after the pruning with the beam width of $N$. However, this standard pruning, or \emph{width-pruning}, cannot prevent the tree from growing indefinitely especially when the input speech is very long. This gradually degrades the efficiency of beam search on recent nodes since more and more hypotheses would be wasted to maintain the old part of the lattice that is already out of the context range of the RNN LMs.

To remedy this issue, we propose an additional pruning method called \emph{depth-pruning}. The procedure is as follows. First, find the $M$-th ancestor of the node with the best hypothesis, where $M$ is the beam depth. Then, the ancestor node becomes a new root node. The pruning is performed by removing the nodes that are not descendants of the new root node. In this way, a beam can be better utilized for recent hypotheses rather than older ones. \figurename~\ref{fig:pruning} shows an example of depth-pruning with the beam depth of 2. Note that the depth of some nodes can be larger than the beam depth. In the following experiments, depth-pruning is performed every 20 frames.

\section{Experiments}
\label{sec:evaluation}

\begin{figure}[t]
\centering
\centerline{%
\begin{tikzpicture}
\begin{axis}
[
width=\columnwidth,
height=0.6\columnwidth,
compat=1.3,
xmin=0.0,
ymin=8,
xmax=100,
ymax=12,
label style={font=\footnotesize},
xlabel={Beam depth (characters)},
ylabel={WER (\%)},
xlabel shift=-2pt,
ylabel shift=-2pt,
legend columns=2,
legend style={
	font=\scriptsize,at={(0.97,0.95)},anchor=north east,
	/tikz/column 2/.style={
	column sep=5pt}
},
tick label style={font=\scriptsize},
ymajorgrids,
minor x tick num=4,
minor y tick num=4,
log basis x={10},
xtick pos=both,
xtick align=inside,
major tick style={line width=0.010cm, black},
major tick length=0.10cm
]%
\legend{{SI-284, BW=128},{SI-284, BW=512},{SI-ALL, BW=128},{SI-ALL, BW=512}}
\addplot[color=black, solid, mark=square, mark size=1.5, mark repeat=1,mark options=solid]
file{data/online_SI284_BW128.txt};
\addplot[color=orange, solid, mark=triangle, mark size=2, mark repeat=1,mark options=solid]
file{data/online_SI284_BW512.txt};
\addplot[color=cyan, solid, mark=o, mark size=1.5, mark repeat=1,mark options=solid]
file{data/online_SIALL_BW128.txt};
\addplot[color=magenta, solid, mark=x, mark size=2, mark repeat=1,mark options=solid]
file{data/online_SIALL_BW512.txt};
\end{axis}%
\end{tikzpicture}%
}%
\caption{WER of the proposed online decoding on the evaluation set with respect to the beam depth. Experiments are conducted with two acoustic RNNs trained on \texttt{SI-284} and \texttt{SI-ALL} and beam search is performed with the beam width (BW) of 128 and 512.}
\label{fig:online}
\end{figure}
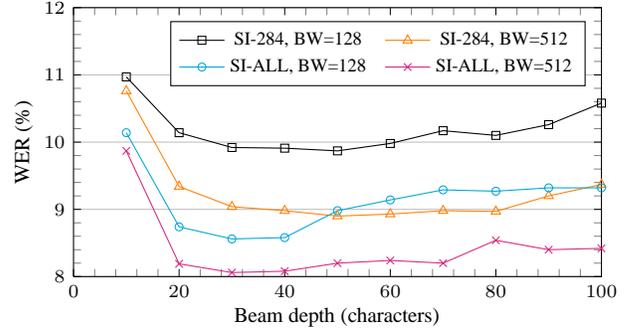

\begin{table}[!t]
\renewcommand{\arraystretch}{1.0}
\caption{CER / WER in percent on the evaluation set with online depth-pruning and offline sentence-wise decoding. The error rates are reported with two acoustic RNNs trained on \texttt{SI-284} (71~hrs) and \texttt{SI-ALL} (167~hrs).}
\label{tbl:method}
\centering
\vspace{2 mm}
\begin{tabular}{l c c c}
\hline
Method & Beam width & \texttt{SI-284} & \texttt{SI-ALL} \\
\hline\hline
Online (no LM) & 512 & 10.96 / 38.37 & \hphantom{0}9.66 / 35.44\\
Online & 128 & 4.25 / 9.87 & 3.56 / 8.56\\
Online & 512 & 3.80 / \bf{8.90} & 3.39 / \bf{8.06}\\
Sentence-wise & 128 & \hphantom{0}4.46 / 10.30 & 3.63 / 8.84\\
Sentence-wise & 512 & 4.04 / 9.45 & 3.38 / 8.28\\
\hline
\end{tabular}
\end{table}

\begin{table}[!t]
\renewcommand{\arraystretch}{1.0}
\caption{Comparison of WERs with other end-to-end speech recognizers in the literature. For reference, WERs of phoneme based GMM/DNN-HMM systems are also reported. All systems are trained with \texttt{SI-284} and evaluated on \texttt{eval92}.} 
\label{tbl:baseline}
\centering
\vspace{2 mm}
\begin{tabular}{l  l  l}
\hline
System & Model & WER \\
\hline\hline
Proposed ISR & Uni. CTC + Char. RNN LM & 8.90\% \\
Graves and Jaitly \cite{graves2014towards} & CTC + Trigram (extended) & \hphantom{0}8.7\% \\
Miao \textit{et al}. \cite{miao2015eesen} & CTC + Trigram (extended) & 7.34\% \\
Miao \textit{et al}. \cite{miao2015eesen} & CTC + Trigram & 9.07\% \\
Hannun \textit{et al}. \cite{hannun2014first} & CTC + Bigram & 14.1\% \\
Bahdanau \textit{et al}. \cite{bahdanau2015end} & Encoder-decoder + Trigram & 11.3\% \\
\hline
Woodland \textit{et al}. \cite{woodland1994large}& GMM-HMM + Trigram & 9.46\% \\
Miao \textit{et al}. \cite{miao2015eesen} & DNN-HMM + Trigram & 7.14\% \\
\hline
\end{tabular}
\end{table}

\begin{figure*}[!t]
	\begin{framed}
	\footnotesize
100: \texttt{HE'S\_THE\_}\\
150: \texttt{HE'S\_THE\_ONLY\_GU}\\
200: \texttt{HE'S\_THE\_ONLY\_GUY\_WHO\_COULD\_S}\\
250: \texttt{HE'S\_THE\_ONLY\_GUY\_WHO\_COULD\_SHOW\_UP\_IN\_THE\_}\\
300: \texttt{...IN\_THE\_PLAZA\_I}\\
350: \texttt{...IN\_THE\_PLAZA\_IN\_ROCK\_R}\\
400: \texttt{...IN\_THE\_PLAZA\_IN\_\textbf{DRAW}\_\textbf{RATE}\_OF\_SEVE}\\
450: \texttt{...IN\_THE\_PLAZA\_IN\_DRAW\_RATE\_OF\_SEVENTY\_FIVE\_THO}\\
500: \texttt{...IN\_THE\_PLAZA\_\textbf{AND}\_DRAW\_\textbf{CROWD}\_OF\_SEVENTY\_FIVE\_THOUSAND\_\textbf{PEO}}\\
550: \texttt{...IN\_THE\_PLAZA\_AND\_DRAW\_CROWD\_OF\_SEVENTY\_FIVE\_THOUSAND\_PEOPLE\_S}\\
600: \texttt{...IN\_THE\_PLAZA\_AND\_DRAW\_CROWD\_OF\_SEVENTY\_FIVE\_THOUSAND\_PEOPLE\_SAYS\_ONE\_LA}\\
650: \texttt{...IN\_THE\_PLAZA\_AND\_DRAW\_CROWD\_OF\_SEVENTY\_FIVE\_THOUSAND\_PEOPLE\_SAYS\_ONE\_LATIN\_DIPLOM}\\
700: \texttt{...IN\_THE\_PLAZA\_AND\_DRAW\_CROWD\_OF\_SEVENTY\_FIVE\_THOUSAND\_PEOPLE\_SAYS\_ONE\_LATIN\_DIPLOMAT} \vspace{3pt} \\
Ground truth: \texttt{HE'S\_THE\_ONLY\_GUY\_WHO\_COULD\_SHOW\_UP\_IN\_THE\_PLAZA\_AND\_DRAW\_}\\
\hphantom{Ground truth: }\texttt{\textbf{A\_}CROWD\_OF\_SEVENTY\_FIVE\_THOUSAND\_PEOPLE\_SAYS\_ONE\_LATIN\_DIPLOMAT}
	\end{framed}
	\caption{Example of ISR partial results. The best hypothesis is shown at every 50 frames (500 ms). The word ``ROCK" is corrected to ``DRAW" after hearing ``RATE" and ``IN DRAW RATE" to ``AND DRAW CROWD" while hearing ``PEOPLE".}
	\label{fig:incremental}
\end{figure*}

The proposed ISR system is evaluated on a single 42-minute speech stream that is formed by concatenating all 333 utterances in the evaluation set, \texttt{eval92} (WSJ Nov'92 20k evaluation set). We use $\alpha=2.0$ and $\beta=1.5$ for the system trained with \texttt{SI-284}, and $\alpha=1.5$ and $\beta=2.0$ for the other one trained with \texttt{SI-ALL}.

The effects of beam depth and width to the final WER are examined in \figurename~\ref{fig:online}. The gap between the beam width of 128 and 512 is roughly 0.5\% to 1\% WER. However, there was little difference when the beam width increases from 512 to 2048 in our preliminary experiments.
The best performing beam depths are 50 and 30 for the \texttt{SI-284} and \texttt{SI-ALL} systems, respectively. This means the \texttt{SI-ALL} system can recognize speech more immediately than the \texttt{SI-284} system. We consider this is because the acoustic model of the \texttt{SI-ALL} system can embed stronger language model due to increased training data, and can make decision more precisely without relying on the external language model much. The character error rate (CER) and WER are reported in \tablename~\ref{tbl:method} with the optimal beam depths. For comparison, we also report sentence-wise offline decoding results without depth-pruning.

The proposed ISR system is compared with other end-to-end word-level speech recognition systems in \tablename~\ref{tbl:baseline}. The other systems perform sentence-wise offline decoding with bidirectional RNNs. The best result was achieved by Miao \textit{et al}. \cite{miao2015eesen} with a CTC-trained deep bidirectional LSTM network and a retrained trigram LM with extended vocabulary. The systems with the original trigram model provided with the WSJ corpus perform worse than our ISR system with character-level RNN LM. On the other hand, our system is beaten by the other ones with extended trigram models. However, more precise comparison of the decoding stages should be done by employing the same CTC model.

\figurename~\ref{fig:incremental} shows the incremental speech recognition result with the proposed ISR system. The best hypothesis is reported every 50 frames (500 ms). It is shown that the past best result can be corrected by making use of the additional speech input. For example, the word ``ROCK'' is changed to ``DRAW'' in the frame 450 by listening the word ``RATE''. Moreover, the correction of ``IN DRAW RATE'' to ``AND DRAW CROWD'' during hearing the word ``PEOPLE'' in the frame 500 is a good evidence that long term context can also be considered.

\section{Concluding remarks}
\label{sec:conclusion}

A character-level incremental speech recognizer is proposed and analyzed throughout the paper. The proposed system combines a CTC-trained RNN with a character-level RNN LM through tree-based beam search decoding. For online decoding with very long input speech, depth-pruning is proposed to prevent indefinite growth of the search tree. When the proposed model is trained with WSJ \texttt{SI-284}, 8.90\% WER can be achieved on the very long speech that is formed by concatenating all utterances in the WSJ \texttt{eval92} evaluation set. The incremental recognition result shows the evidence that character-level RNN LM can learn dependencies between two words even when they are five words apart, which are hard to be caught using conventional $n$-gram back-off language models.

Note that the proposed system only requires speech and text corpus for training. External lexicon or senone modeling is not needed for training, which is a huge advantage. Moreover, it is expected that OOV words or infrequent words such as names of places or people can be dictated as they are pronounced. 

%

\pagebreak

\bibliographystyle{IEEEbib}
\bibliography{refs}

\end{document}